\ifdefined\pdfoutput
\pdfoutput=1
\fi
\documentclass[conference]{IEEEtran}
\IEEEoverridecommandlockouts

\usepackage{enumitem}

\usepackage[utf8]{inputenc} % allow utf-8 input
\usepackage[T1]{fontenc}    % use 8-bit T1 fonts
\usepackage[scaled=0.85]{beramono}
\usepackage{hyperref}       % hyperlinks
\usepackage{url}            % simple URL typesetting
\usepackage{booktabs}       % professional-quality tables
\usepackage{amsfonts}       % blackboard math symbols
\usepackage{nicefrac}       % compact symbols for 1/2, etc.
\usepackage{microtype}      % microtypography
\usepackage{xcolor}         % colors
\usepackage{graphicx}
\usepackage{multirow}
\usepackage{subfig}      % subfigures
\usepackage{colortbl}    % table cell colors
\usepackage{xcolor}      % text and symbol colors
\usepackage{pifont}      % check and cross symbols
\usepackage{adjustbox}
\usepackage{float}
\usepackage{balance}
\definecolor{mygreen}{rgb}{0,0.6,0}
\definecolor{myred}{rgb}{0.8,0,0}

\definecolor{cycle2}{RGB}{106, 191, 0}
\definecolor{cycle3}{RGB}{191, 0, 0}
\definecolor{amber}{rgb}{1.0, 0.75, 0.0}

\newcommand{\cmark}{\textcolor{cycle2}{\ding{52}}} %
\newcommand{\xmark}{\textcolor{cycle3}{\ding{56}}}

\newcommand{\nop}[1]{}
\newcommand{\partitle}[1]{\medskip \noindent \textbf{#1.}}
\newcommand{\noindentparagraph}[1]{\medskip \noindent \textbf{#1} }

\begin{document}

\title{[Experiment, Analysis, and Benchmark] DaDaDa: A Dataset for Data Pricing in Data Marketplaces}

\author{
\IEEEauthorblockN{
Qiheng Sun\textsuperscript{1,2},
Hongwei Zhang\textsuperscript{2},
Junxu Liu\textsuperscript{1},
Xiaokai Mao\textsuperscript{2},
Jinfei Liu\textsuperscript{2},
Kui Ren\textsuperscript{2},
Haibo Hu\textsuperscript{1}
}
\IEEEauthorblockA{
\textsuperscript{1}The Hong Kong Polytechnic University, Hong Kong, China\\
\textsuperscript{2}Zhejiang University, Hangzhou, China
}
\IEEEauthorblockA{
\{qiheng.sun, junxu.liu, haibo.hu\}@polyu.edu.hk\\
\{hongweizhang, xiaokaimao, jinfeiliu, kuiren\}@zju.edu.cn
}
}

\maketitle

\begin{abstract}
\label{sec:abstract}
High-quality data drives machine learning advances across industries. Recognizing the value of data, data transactions are increasingly common, giving rise to many data marketplaces, e.g., AWS Marketplace, Databricks, and Datarade. However, determining the appropriate prices for data products remains a significant challenge due to the unique properties of data products. Traditional pricing methods in economics can be categorized into the cost approach, the income approach, and the sales comparison approach. The cost approach fails in data pricing due to near-zero marginal cost from data replication, and the income approach fails due to inherently unpredictable data revenue. The sales comparison approach remains viable, yet its application is hindered by the absence of standardized pricing benchmarks for data products across marketplaces. To address this challenge, we introduce \texttt{DaDaDa}, the first dataset for data product pricing, containing metadata for 16,147 data products from 9 major data marketplaces worldwide. \texttt{DaDaDa} enables the training of pricing models, thereby establishing price benchmarks for new data products. In addition, \texttt{DaDaDa} can be utilized for other important tasks in data markets, such as data product classification and retrieval. Experiments and a retrieval prototype demonstrate the effectiveness of \texttt{DaDaDa} for pricing, classification, and retrieval of data products. The dataset and code are available at \url{https://github.com/ZJU-DIVER/DaDaDa}.
\end{abstract}


\section{Introduction}
\label{sec:introduction}
The immense value created by data-driven machine learning models stems from accessible, diverse, and well-structured data~\cite{chai2022data}. However, publicly available data are running out, as noted in recent discussions of AI data scarcity~\cite{jones2024ai}. Moreover, such data often lack domain specificity and structural consistency. In contrast, private data sources not only retain vast amounts of unused data but also provide curated, task-specific datasets. This gap between the need for high-quality data and the limitations of public sources drives the growth of private data transactions~\cite{fernandez13data}, giving rise to dedicated marketplaces such as AWS Marketplace~\cite{AWS}, Databricks~\cite{Databricks}, and Datarade~\cite{Datarade}.  These marketplaces provide a convenient avenue for showcasing data products and facilitate connections between data sellers and buyers. 

A critical function of marketplaces is enabling pricing for products, which determines transaction viability and market liquidity~\cite{pei2023data}. There are three commonly used pricing methods: (1) the cost approach, (2) the income approach, and (3) the sales comparison approach~\cite{baum2021property, damodaran2012investment, dolan1995you}. The cost approach sets the price of a product by assessing the cost of replacing or reproducing it. Evaluating the cost of data products is difficult due to the fact that they can be replicated and distributed at almost zero cost. The income approach determines the price of a product based on the income that it is expected to bring. Predicting the income of data products is challenging because they can generate different revenues across use cases and be sold multiple times to distinct customers. Due to inherent issues with these two methods in the context of data products, the sales comparison approach is more applicable and widely adopted in data markets. The sales comparison approach utilizes existing pricing information of similar products to estimate the price of the commodity to be sold. This reduces information asymmetry, fosters mutual acceptance between buyers and sellers, and ultimately facilitates transaction completion in data markets.

With the rapid development of data marketplaces, the sales comparison approach holds significant potential for proper data product pricing by leveraging aggregated pricing signals across platforms. However, its practical implementation faces a critical barrier: comparing and evaluating data products across marketplaces is challenging due to different taxonomies and inconsistent metadata granularity. This complexity is further compounded by the limited capacity of pricing agents to conduct exhaustive cross-marketplace comparisons, as the cognitive burden and time required to manually align disparate product specifications often exceed practical constraints. This structural fragmentation limits the applicability of the sales comparison approach despite its theoretical advantages, thereby constraining market efficiency through suboptimal pricing decisions and reduced transaction volumes. Consequently, enabling automated cross-marketplace price benchmarking represents a pivotal challenge for realizing the full potential of data markets.

To enhance the applicability of the sales comparison approach in data markets, we propose \texttt{DaDaDa} (A \underline{\texttt{Da}}taset for \underline{\texttt{Da}}ta Pricing in \underline{\texttt{Da}}ta Marketplaces), containing metadata for 16,147 data products from 9 major data marketplaces and providing price references for pricing decisions. We investigate 60 data marketplaces and carefully select 9 of them based on specific criteria, including the accessibility of their data product catalogs, the diversity of data products they offer, and the comprehensiveness of the metadata they provide. This careful selection process ensures that \texttt{DaDaDa} offers a diverse and representative sample of a broad data marketplace ecosystem. To create \texttt{DaDaDa}, we use advanced web crawling techniques to extract detailed metadata for each data product, then create fields such as title, description, provider, price, volume, size, and category. Following metadata scraping, we conduct thorough data preprocessing to unify various metadata fields into a consistent format, making it easy to analyze and use for various applications. The resulting \texttt{DaDaDa} dataset serves as a unified, analyzable benchmark, making the systematic research and application of the sales comparison approach for data product pricing feasible. 

We conduct experiments to demonstrate the effectiveness of using \texttt{DaDaDa} for essential tasks in data markets.
\begin{itemize}[leftmargin=*]
    % \item \textit{Data product pricing.} This task involves pricing data products based on their metadata. We train several regression models on \texttt{DaDaDa} to predict the prices of data products. The best-performing model achieved an $R^2$ score of 83\%.
    % \item \textit{Data product classification.} This task involves classifying data products into predefined categories. We fine-tune two pretrained multilingual models to extract semantic information from the ``title'' and ``descriptions'' fields in \texttt{DaDaDa}, achieving over 82\% classification accuracy.
    \item \textit{Data product pricing.} This task involves pricing data products based on their metadata. We train several regression models on \texttt{DaDaDa} to predict product prices, providing empirical baselines for future research on automated data pricing.
    \item \textit{Data product classification.} This task involves classifying data products into predefined categories. We fine-tune pretrained multilingual models using product titles and descriptions, showing that \texttt{DaDaDa} supports semantic analysis and category prediction in data marketplaces.
    \item \textit{Data product retrieval.} This task involves retrieving relevant data products from a large collection of data products in response to user queries. We use Elasticsearch~\cite{ElasticSearch} to develop a search engine within \texttt{DaDaDa}, allowing users to quickly find interesting data products along with corresponding pricing information.
\end{itemize}

The rest of the paper is organized as follows. We first review related work on data product pricing in Section~\ref{section:related_work}. Section~\ref{section:dataset_overview} provides an overview of the \texttt{DaDaDa} dataset. 
The dataset construction pipeline is detailed in Section~\ref{section:dataset_pipeline}. Section~\ref{section:applications} explores the various applications of \texttt{DaDaDa}. Finally, we conclude this paper and discuss the limitations and future work in Section~\ref{section:conclusion}.

\section{Related Work}
\label{section:related_work}

\subsection{Theoretical Data Pricing Mechanisms} 

Existing theoretical studies on data pricing can be broadly organized around three types of mechanisms. The first line studies how to price data access and query answers under formal economic constraints. Query-based pricing work establishes arbitrage-free and discount-free prices for database queries, and further develops scalable query-pricing systems and revenue-maximization models~\cite{koutris2015query,DBLP:conf/sigmod/DeepK17,DBLP:journals/pvldb/ChawlaDKT19}. Closely related, Li et al.~\cite{li2014theory} develop a theoretical framework for pricing noisy query responses with compensation mechanisms based on privacy loss.

A second line designs market mechanisms for acquiring, selling, or allocating data when participants have private information. Price-quantity mechanisms account for buyers' private valuations of data subsets~\cite{DBLP:journals/isr/MehtaDJM21}, while auction- and Shapley-value-based marketplace designs aim to elicit truthful bids and ensure fair revenue distribution among data owners~\cite{agarwal2019marketplace}. Peer-prediction mechanisms further support truthful data acquisition when no test dataset is available~\cite{DBLP:conf/nips/ChenSZ20}. Related procurement work considers scenarios where individuals incur privacy costs, using truthful auctions under differential privacy to purchase private data subject to budget or accuracy constraints~\cite{ghosh2011selling}.

A third line extends pricing theory from raw data to machine-learning artifacts and privacy-aware marketplaces. Model-pricing frameworks prevent arbitrage while balancing seller revenue and buyer affordability~\cite{DBLP:conf/sigmod/ChenK019}, and end-to-end model marketplaces incorporate differential privacy into compensation and pricing functions to support fairness, privacy preservation, and revenue maximization~\cite{liu2021dealer}. Despite their theoretical rigor, these approaches share a critical limitation: they lack empirical validation using real-world pricing data from operational data marketplaces.  
%The absence of comprehensive, real-world pricing benchmarks has hindered both the practical application and empirical evaluation of these theoretical frameworks.

\subsection{Empirical and Commercial Studies} Complementing theoretical work, empirical studies examine commercial data marketplaces from an empirical perspective. Azcoitia et al.~\cite{azcoitia2022survey} conduct a comprehensive survey of commercial data trading entities, analyzing data types, business models, and underlying technologies. In subsequent work, they investigate specific features that influence data product pricing~\cite{azcoitia2023understanding}. Hanspach et al.~\cite{hanspach2024algorithms} and Wieting and Sapi~\cite{wieting2021algorithms} conduct empirical analyses of pricing on major e-commerce platforms, analyzing pricing data from thousands of products to understand automated pricing behavior in digital marketplaces. Azcoitia et al.~\cite{azcoitia2021price} analyze price distributions and price-driving features, highlighting both the feasibility and the challenges of cross-marketplace comparisons. However, these studies primarily focus on descriptive analysis rather than providing structured datasets for algorithmic development and benchmarking.

In summary, existing research suffers from a critical disconnect in that theoretical pricing mechanisms remain unvalidated due to the absence of real-world pricing benchmarks, while empirical marketplace studies lack structured datasets needed for algorithmic development. %\texttt{DaDaDa} bridges this gap by providing the first comprehensive dataset that enables both empirical validation of pricing theories and development of practical marketplace algorithms through harmonized cross-platform metadata.

\section{Dataset Overview}
\label{section:dataset_overview}

\begin{figure}[htbp]
    \centering
    \begin{minipage}[b]{0.46\textwidth}
        \centering
        \includegraphics[width=\textwidth]{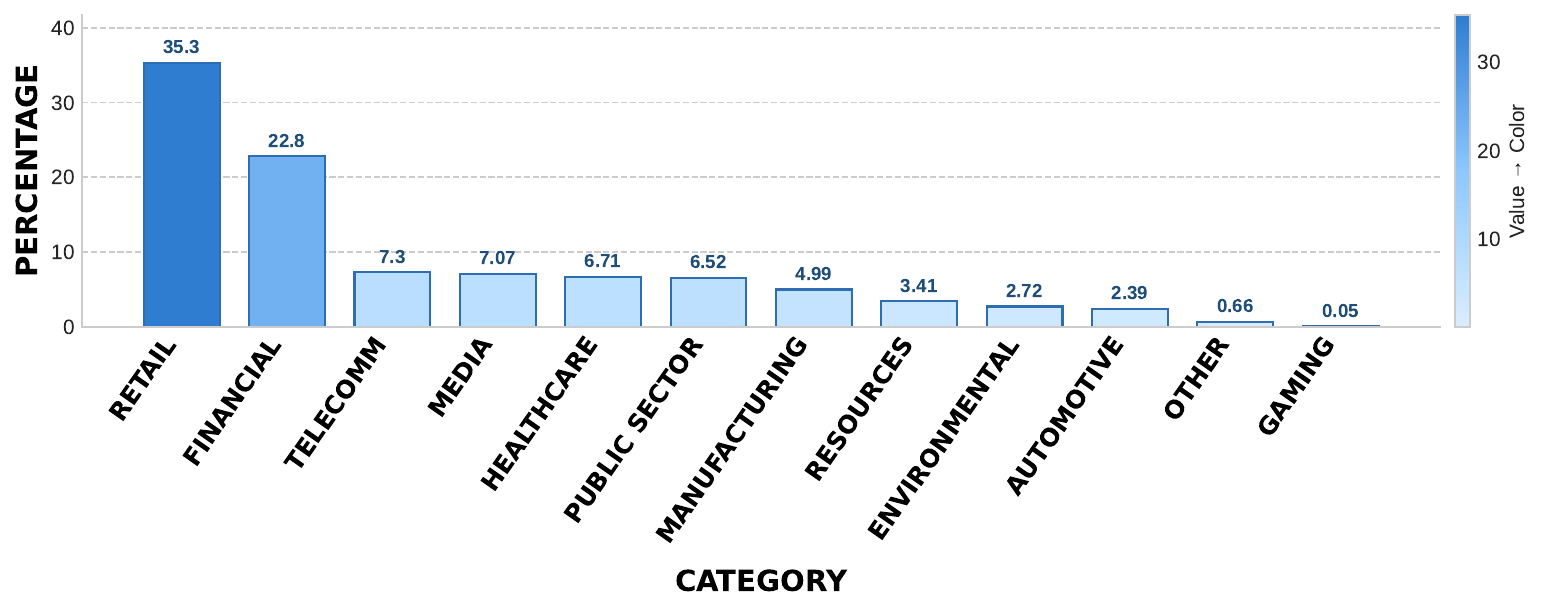}
        \caption{Percentage of each category in \texttt{category}.}
        \label{fig:category}
    \end{minipage}
    \hfill
    \begin{minipage}[b]{0.46\textwidth}
        \centering
    \includegraphics[width=\textwidth]{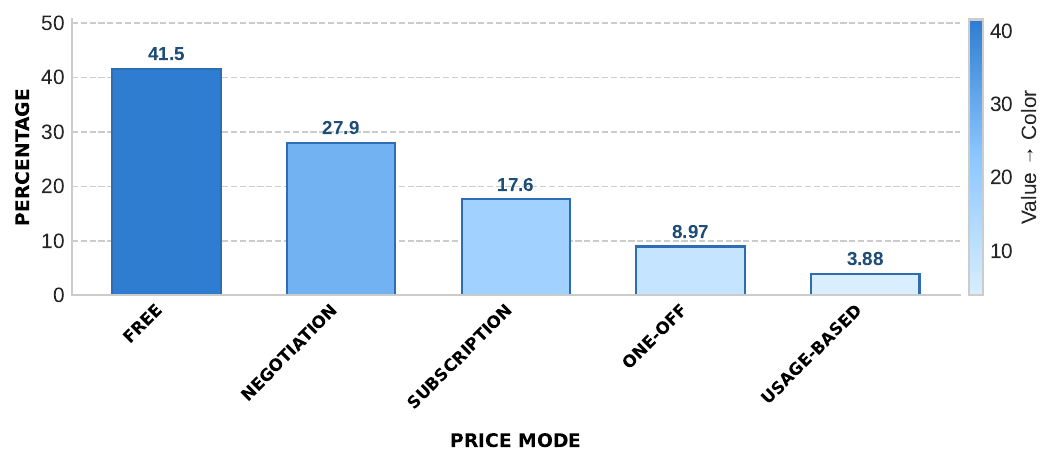}
        \caption{Percentage of each pricing mode in \texttt{price\_mode}.}
        \label{fig:mode}
    \end{minipage}
\end{figure}

In this section, we present an overview of \texttt{DaDaDa}. A detailed description of the construction pipeline is provided in Section \ref{section:dataset_pipeline}.

Overall, \texttt{DaDaDa} distinguishes itself from existing benchmark datasets through three key contributions. (1) \textbf{Bridging the Gap.} It is the first real-world dataset for data product pricing, enabling both empirical validation of pricing theories and the development of practical marketplace algorithms. (2) \textbf{Extensive Coverage.} It provides detailed metadata for 16,147 data products sourced from 9 leading data marketplaces worldwide, facilitating in-depth analysis of the commercial data landscape. (see Section \ref{section:data_marketplace_selection} for the full list).
(3) \textbf{Standardized Metadata.} Each data product is described using a standardized metadata template, curated through web crawling, scraping, and thorough preprocessing to capture a comprehensive set of relevant attributes. The complete set of metadata attributes is detailed below.

% The attributes comprising \texttt{DaDaDa} are detailed below.
% The dataset can facilitate analysis of the commercial data landscape, including data product pricing, categorization, and retrieval. 

\textbullet\ \texttt{Title.} The title of the data product.

\textbullet\ {\texttt{Platform.}} The marketplace name hosting the data product.

\textbullet\ {\texttt{Provider.}} The name of the data provider as listed on the data marketplace. The released table contains 1,991 named data providers. Among them, ``Techsalerator'' is the leading provider, offering a total of 644 data products, making it the supplier with the highest volume of offerings on the market. For products whose source platform does not disclose the provider identity, we preserve an empty provider value rather than imputing a potentially incorrect name.

\textbullet\ {\texttt{Description.}} The detailed description of the data product.

\textbullet\ {\texttt{Volume.}} The number of records.

\textbullet\ {\texttt{Size.}} The data size (in Byte) provided by the data product.

\textbullet\ {\texttt{Dimension.}} The number of data features. 

\textbullet\ {\texttt{Coverage.}} The countries covered by the data product. 

\textbullet\ {\texttt{Update\_frequency.}} The frequency between data product updates as announced by the seller, such as ``monthly'', ``daily'', and ``real-time''. Most data products adopt ``no-update'' and ``daily''.

\textbullet\ {\texttt{Data\_sample.}}
The filename of the data sample if available. We download and store the data sample of data products in an additional folder.

\textbullet\ {\texttt{Category.}} 
The original category of data product may vary across different data marketplaces, each with its own way of categorization. We align the data categories from other marketplaces with the AWS Marketplace categories through manual labeling.
As shown in Figure \ref{fig:category}, Retail (Retail, Location, and Marketing Data) and Financial (Financial Services Data) have the highest proportions.

\textbullet\ {\texttt{Price\_mode.}} The pricing mode of the data product. There are five pricing modes: (1) {negotiation} mode where data buyers need to negotiate the price with data providers, (2) {free} mode where the data is provided at no cost, (3) {subscription} mode where data buyers are charged a recurring fee on a monthly or annual basis, (4) {one-off} mode where data buyers pay a one-time fee to access the data permanently, and (5) {usage-based} mode where data buyers are charged based on the amount of data they consume, such as the volume of data downloaded or the number of API calls. As shown in Figure~\ref{fig:mode}, most data products in the data marketplaces are either free or priced through negotiation. A smaller proportion (subscription, one-off, or usage-based) has clear prices.

\textbullet\ {\texttt{Price.}} We use USD (\textdollar) as the currency unit. If the pricing mode is free, the price is set to 0. Since data products with negotiation mode do not have a clear price, we exclude them from the data product pricing task in Section~\ref{sec:pricing}. As a result, negotiation-mode data products are assigned a price of 0. If the pricing mode is subscription, the price represents the subscription cost for 12 months. If the pricing mode is usage-based, the price reflects the cost for a single usage. These two standardizations facilitate price comparisons within the same mode. We do not directly compare data products across different modes, as the price mode is the most significant factor influencing the price of a data product when training the pricing model.

\textbullet\ {\texttt{Url.}} The web address of the detail page of the data product.

\noindentparagraph{Dataset Statistics and Release.}
The released dataset contains 16,147 data products described by 14 standardized metadata fields. The products come from nine marketplaces with different geographic focuses and platform designs: Datarade (4,396 products), AWS Data Exchange (3,702), Snowflake (2,585), Databricks (2,004), Zhejiang Big Data Exchange (838), Canton Data Exchange (792), Beijing International Data Exchange (714), Guiyang Global Big Data Exchange (669), and Shanghai Data Exchange (447). This coverage is important for pricing research because it includes both international cloud marketplaces and regional data exchanges, whose metadata conventions and pricing disclosure practices differ substantially. In addition to the final table, we release the raw crawled files, intermediate preprocessed files, preprocessing scripts, crawler implementations, downstream experiment code, and 255 data product samples when samples are publicly available from the source platforms. These artifacts allow users to inspect how each product moves from raw marketplace metadata to the unified \texttt{DaDaDa} schema.

\noindentparagraph{Price Coverage.}
Some products provide explicit nonzero prices under subscription, one-off, or usage-based modes; these products constitute the labeled subset used in the data product pricing task. The remaining products are free or negotiation-based. We retain them in the complete dataset because they are still useful for studying marketplace composition, category classification, retrieval, and pricing-mode prediction. This design follows the practical structure of current data marketplaces: many sellers intentionally hide exact prices and require buyers to negotiate, while buyers still need comparable metadata to identify alternatives and understand market positioning. By separating \texttt{price\_mode} from \texttt{price}, \texttt{DaDaDa} supports both direct price regression on products with explicit prices and broader analysis of disclosure patterns across platforms.

\noindentparagraph{Marketplace Coverage and Benchmark Scope.}
Table~\ref{tab:platform_profile} summarizes the platform-level coverage of \texttt{DaDaDa}, including product counts, pricing-mode composition, and public sample availability. The purpose of this profiling table is to make the released snapshot auditable: users can identify which marketplaces contribute products to the unified schema and choose task-specific subsets according to their experimental needs.

This platform-level profile documents the scope of the released snapshot rather than defining a single downstream task. Since public data marketplaces organize listings using different pricing and disclosure conventions, the table helps users understand the composition of \texttt{DaDaDa} and select task-specific subsets for pricing, classification, retrieval, or marketplace analysis.
The same principle applies to structured metadata. Attributes such as \texttt{volume}, \texttt{size}, and \texttt{dimension} are used when they are reliably disclosed, while textual fields such as \texttt{title} and \texttt{description} provide a consistent semantic representation across platforms. This design preserves broad marketplace coverage and allows downstream models to combine textual and structured signals according to the requirements of each task.

\begin{table*}[t]
\begingroup
\small
\centering
\caption{Platform-level profiling of \texttt{DaDaDa}. ``Explicit'' denotes products with explicit nonzero prices. ``Free'', ``Neg.'', ``Sub.'', ``One-off'', and ``Usage'' report the numbers of products under the free, negotiation, subscription, one-off, and usage-based pricing modes, respectively. ``Public samples'' denotes publicly available data sample files downloaded from product pages.}
\label{tab:platform_profile}
\adjustbox{width=\textwidth}{
\begin{tabular}{lrrrrrrrrr}
\toprule
Marketplace & Products & Explicit & Explicit (\%) & Free & Neg. & Sub. & One-off & Usage & Public samples \\
\midrule
Datarade & 4,396 & 1,986 & 45.2 & 0 & 2,410 & 1,082 & 695 & 209 & 255 \\
AWS Data Exchange & 3,702 & 1,606 & 43.4 & 2,096 & 0 & 1,592 & 0 & 14 & 0 \\
Snowflake & 2,585 & 237 & 9.2 & 2,348 & 0 & 172 & 12 & 53 & 0 \\
Databricks & 2,004 & 0 & 0.0 & 1,610 & 394 & 0 & 0 & 0 & 0 \\
Zhejiang Big Data Exchange & 838 & 833 & 99.4 & 0 & 5 & 1 & 735 & 97 & 0 \\
Canton Data Exchange & 792 & 14 & 1.8 & 0 & 778 & 7 & 2 & 5 & 0 \\
Beijing International Data Exchange & 714 & 202 & 28.3 & 510 & 0 & 0 & 6 & 198 & 0 \\
Guiyang Global Big Data Exchange & 669 & 51 & 7.6 & 142 & 476 & 0 & 0 & 51 & 0 \\
Shanghai Data Exchange & 447 & 0 & 0.0 & 0 & 447 & 0 & 0 & 0 & 0 \\
\midrule
All & 16,147 & 4,929 & 30.5 & 6,706 & 4,510 & 2,854 & 1,450 & 627 & 255 \\
\bottomrule
\end{tabular}
}
\endgroup
\end{table*}

\noindentparagraph{Metadata Sparsity.}
The dataset also preserves the incompleteness of real marketplace metadata. Fields such as \texttt{title}, \texttt{description}, \texttt{platform}, \texttt{category}, \texttt{price\_mode}, and \texttt{url} are complete after preprocessing, while numerical fields such as \texttt{volume}, \texttt{size}, and \texttt{dimension} are disclosed unevenly across platforms. Instead of discarding products from metadata-poor marketplaces, we use explicit default values when a field cannot be reliably inferred from the product page. This choice makes the benchmark more representative of real data marketplace search and pricing conditions, where buyers often rely on textual descriptions and partial structured metadata. It also explains why the textual representation of \texttt{title} and \texttt{description} plays a major role in the pricing experiments in Section~\ref{sec:pricing}.

\noindentparagraph{Benchmark Implications.}
Structured numerical attributes in data marketplaces are not disclosed as consistently as textual fields. When fields such as \texttt{volume}, \texttt{size}, and \texttt{dimension} cannot be reliably extracted from the original product pages, we preserve explicit default values rather than inventing product properties that sellers do not publish. The implication is that real data-product metadata is largely text-centric: \texttt{title} and \texttt{description} are the most consistently available high-content fields across marketplaces, while numerical attributes are useful complementary signals only when sellers or platforms expose them reliably.

This observation directly shapes the benchmark design of \texttt{DaDaDa}. First, the pricing task should combine semantic representations with structured features instead of relying on purely tabular metadata, because a structured-only model would either discard many products or depend heavily on artificial imputation. Second, the classification task is naturally formulated as a multilingual text understanding problem, since category evidence is primarily expressed in product titles and descriptions. Third, the retrieval task should combine full-text search with optional structured filters: buyers can search semantically over all products, while using price, coverage, update frequency, size, volume, or dimension filters only when those fields are available. The SHAP analysis in Section~\ref{sec:pricing} further supports this design choice: \texttt{description} is the most influential feature group across subscription, one-off, and usage-based pricing modes, while structured attributes such as coverage, update frequency, and dimension provide mode-specific signals when available.

\begin{figure}[h]
    \centering
    \includegraphics[width=1\linewidth]{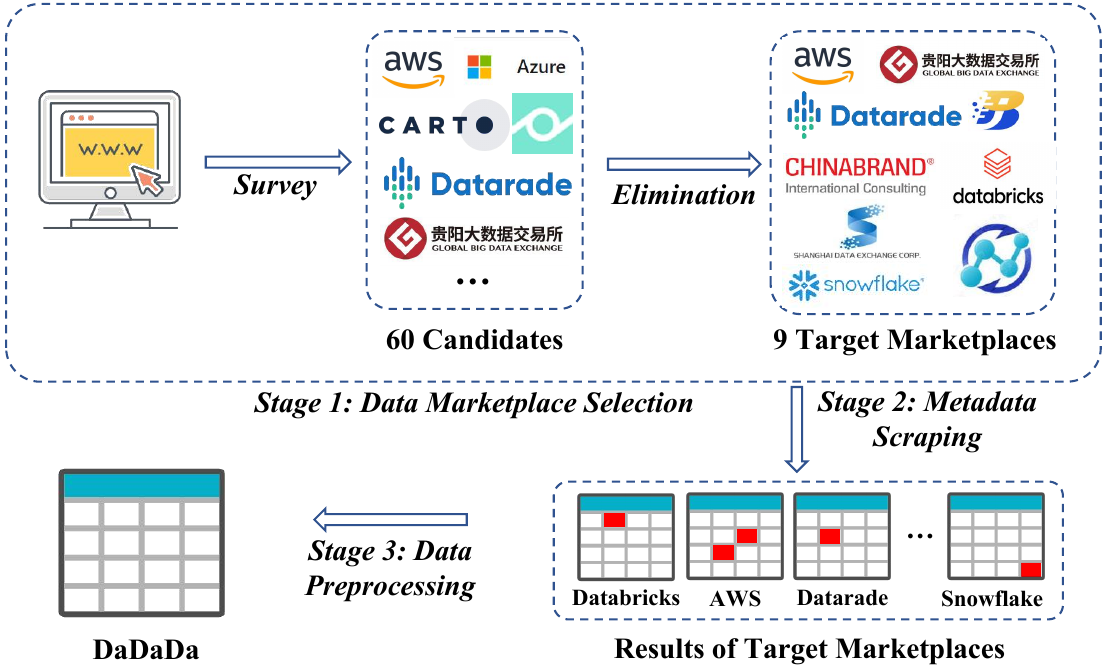}
    \caption{Overview of the data collection pipeline. First, we identify and select 9 major data marketplaces that meet our criteria. Next, we use web scraping techniques to extract metadata from the selected data marketplaces. Finally, we carry out data preprocessing, removing errors or inconsistencies, combining data from different sources, and converting the data into a unified format for analysis.}
    \label{fig:pipeline}
\end{figure}

\section{Construction Pipeline}
\label{section:dataset_pipeline}
We develop a comprehensive data collection pipeline to gather and organize data from various data marketplaces. As shown in Figure \ref{fig:pipeline}, this pipeline consists of three main stages: (1) data marketplace selection, (2) metadata scraping, and (3) data preprocessing. The details of each stage are described as follows.
 
\subsection{Data Marketplace Selection} 
\label{section:data_marketplace_selection}
% The data marketplace selection is driven by a rigorous set of criteria designed to ensure the quality and robustness of the raw data sources. These criteria include:

Ensuring that raw data sources satisfy rigorous quality and robustness standards is fundamental to developing reliable pricing benchmarks.
To this end, we first identify candidate data marketplaces through a selection process guided by three criteria:
\begin{enumerate}[leftmargin=*, nosep]
\item \textit{Data Accessibility.} Data marketplaces must provide open access to their data product catalogs and permit data extraction through web crawling techniques. % The data marketplace grants access to its data product catalogs for free and allows the use of web crawlers for data extraction.
\item \textit{Product Variety.} Data marketplaces must encompass a sufficient range of product categories to ensure diversity. In our framework, we adopt a practical threshold of eight categories and manually assess coverage using AWS Marketplace classification criteria as the benchmark. % We use the classification criteria of AWS Marketplace as a benchmark and manually evaluated the category coverage of each data marketplace. To ensure variety, the selected marketplaces must include at least 8 data product categories. 
\item \textit{Metadata Comprehensiveness.} %We defined a metadata template with fields including but not limited to \texttt{title}, \texttt{description}, \texttt{category}, \texttt{provider}, \texttt{price}, \texttt{volume}, \texttt{size}, \texttt{dimension}, and \texttt{coverage}. 
Data marketplaces must include a substantial portion of the essential metadata fields associated with each data product. Specifically, we require that each marketplace cover at least 70\% of the metadata attributes listed in Section~\ref{section:dataset_overview}. %To evaluate the comprehensiveness of each data marketplace, we assessed the number of metadata fields provided for each data product. 
\end{enumerate}

\noindentparagraph{Selection Rationale.}
The three criteria balance coverage, diversity, and reproducibility. Marketplace size alone is insufficient: a large catalog may be concentrated in a narrow domain or expose minimal metadata, while a smaller marketplace can still contribute structured product pages, multiple categories, and stable public listings. We therefore first require crawlable catalogs and then assess whether each candidate adds diverse and comparable metadata to the unified schema. This also prevents the benchmark from being dominated by a few high-volume but narrow catalogs. This protocol helps \texttt{DaDaDa} represent a broad marketplace ecosystem while remaining reproducible from public sources.

Our selection process starts with an extensive survey of data marketplaces worldwide, identifying 60 initial candidates (see Table~\ref{tb:alldm} for more details). Candidates that do not meet the first criterion are eliminated in the first stage. For example, Azure~\cite{Azure} is removed because it requires payments to access data products. This filtering step yields a shortlist of 20 accessible candidates, as shown in Table~\ref{tb:dm}. Next, we evaluate these candidates against the second and third criteria, gathering information regarding the number of data products and categories available in each marketplace. Our analysis reveals that while some marketplaces offer a large volume of data products, they are often concentrated in specific domains. For example, our manual inspection of representative product samples shows that Advaneo specializes in IoT data and Carto focuses on spatial data. Additionally, data marketplaces like Western China Data Exchange~\cite{Western} and Google Cloud~\cite{Googlecloud} provide only limited product information, typically restricted to titles and brief descriptions. After removing these cases from consideration, we ultimately identify 9 candidates, which are listed in the upper section of Table~\ref{tb:dm}.

\begin{table}[t]

  \caption{Summary of the 60 initial data marketplace candidates. }
  \label{tb:alldm}
\adjustbox{width=\linewidth}{
  \centering
  \begin{tabular}{llc}
    \toprule
    % \multicolumn{2}{c}{Part}                   \\
    % \cmidrule(r){1-2}
    Marketplace     & URL   & Accessibility \\
    \midrule
Advaneo & \url{https://www.advaneo-datamarketplace.de/database/en/} & \xmark \\
Agdatahub & \url{https://agdatahub.eu/en/} & \xmark \\
Aircloak & \url{https://aircloak.com/} & \xmark \\
AMO & \url{https://www.amo.foundation/} & \xmark \\
AWS Marketplace & \url{https://aws.amazon.com/marketplace/} & \cmark \\
Azure & \url{https://azure.microsoft.com/en-us/products/open-datasets/} & \xmark \\
BattleFin & \url{https://www.battlefin.com/} & \xmark \\
Beibu Gulf Data Exchange & \url{https://www.bbgdex.com/} & \cmark \\
Beijing International Data Exchange & \url{https://www.bjidex.com} & \cmark \\
BurstIQ & \url{https://burstiq.com/} & \xmark \\
Canton Data Exchange & \url{https://www.cantonde.com} & \cmark \\
Carto & \url{https://carto.com/} & \cmark \\
Caruso & \url{https://www.caruso-dataplace.com/} & \xmark \\
CleverData & \url{https://1dmc.io/} & \xmark \\
Cognite & \url{https://www.cognite.com/en/} & \xmark \\
Crunchbase & \url{https://www.crunchbase.com/} & \xmark \\
Data Intelligence Hub & \url{https://dih.telekom.com/en/} & \xmark \\
Databricks & \url{https://marketplace.databricks.com/} & \cmark \\
Databroker & \url{https://www.databroker.global/} & \cmark \\
Dataeum & \url{https://www.dataeum.io/} & \xmark \\
Datarade & \url{https://datarade.ai/} & \cmark \\
Datatang & \url{https://www.datatang.com/} & \cmark \\
Datawallet & \url{https://www.datawallet.com/} & \xmark \\
Datum & \url{https://arxiv.org/pdf/2201.04561} & \xmark \\
Dawex & \url{https://www.dawex.com/en/data-exchange-technology/} & \xmark \\
Decentr & \url{https://decentr.net/} & \xmark \\
GeoDB & \url{https://geodb.com/en/} & \xmark \\
Google Cloud & \url{https://console.cloud.google.com/marketplace} & \cmark \\
Guiyang Global Big Data Exchange & \url{https://www.gzdex.com.cn} & \cmark \\
Handshakes & \url{https://www.handshakes.com.sg/data} & \xmark \\
Health Verity & \url{https://healthverity.com/} & \xmark \\
HERE & \url{https://www.here.com/} & \xmark \\
IOTA & \url{https://www.iota.org/} & \xmark \\
JDEX & \url{https://www.jdex.jp/} & \xmark \\
Knoema & \url{https://knoema.com/} & \cmark \\
LESG & \url{https://www.lseg.com/en/data-analytics} & \xmark \\
Liveramp & \url{https://liveramp.com/} & \xmark \\
LonGenesis & \url{https://longenesis.com/} & \xmark \\
Lotame & \url{https://www.lotame.com/} & \xmark \\
Madana & \url{https://www.madana.io/products.html} & \xmark \\
Mobilithek & \url{https://mobilithek.info/} & \cmark \\
Mydex & \url{https://mydex.org/} & \xmark \\
Nasdaq Data Link & \url{https://data.nasdaq.com/} & \cmark \\
Northern China Data Exchange & \url{https://www.datadmz.com/} & \cmark \\
Salesforce & \url{https://www.salesforce.com/products/} & \xmark \\
Shanghai Data Exchange & \url{https://dtxp.chinadep.com} & \cmark \\
Shenzhen Data Exchange & \url{https://www.szdex.com} & \cmark \\
Space Data Marketplace & \url{https://www.space-data-marketplace.eu/en/onboard/} & \xmark \\
Streamr & \url{https://streamr.network/} & \xmark \\
Terbine & \url{https://terbine.com/} & \xmark \\
The Adex & \url{https://theadex.com/} & \xmark \\
TheTradeDesk & \url{https://www.thetradedesk.com/us/} & \xmark \\
Urgently & \url{https://www.geturgently.com/} & \xmark \\
Veracity & \url{https://www.veracity.com/} & \cmark \\
Vetri & \url{https://vetri.global/} & \xmark \\
Weople & \url{https://weople.space/en/} & \xmark \\
Western China Data Exchange & \url{https://www.westdex.com.cn} & \cmark \\
Xignite & \url{https://www.xignite.com/} & \xmark \\
Zenome & \url{https://zenome.io/} & \xmark \\
Zhejiang Big Data Exchange & \url{https://ditm.zjdex.com} & \cmark \\
    \bottomrule
  \end{tabular}
  }
\end{table}

%We conduct an initial investigation on 60 data marketplaces (see details in Appendix \ref{sec:alldm}). Those that do not meet the first criterion are eliminated in the first step. The eliminated data marketplaces require registration or payments to access data products, such as Azure \cite{Azure}. From this, a list of 20 accessible candidates is created. We then examine whether these candidates meet the second and third criteria, gathering information regarding the number of data products and categories they offer. Our findings are summarized in Table \ref{tb:dm}. 
% \emph{Products} refers to the total number of data products in the data marketplace; \emph{Categories} represents the number of data product categories in the data marketplace; \emph{Variety} and \emph{Comprehensiveness} correspond to the second and third criteria, respectively. Although some data marketplaces have a large number of data products, they focus on specific domains, such as Advaneo for IOT data and Carto for spatial data. Data marketplaces like Western China Data Exchange~\cite{Western} and Google Cloud~\cite{Googlecloud} provide limited product information, offering only titles and brief descriptions. Therefore, we exclude these data marketplaces.

\begin{table}
  \caption{Summary of 20 accessible data marketplaces. The columns (from left to right) report the total counts of data products and product categories, followed by indicators of whether the marketplace meets the second and third criteria.}
  \label{tb:dm}
\adjustbox{width=\linewidth}{
  \centering
  \begin{tabular}{lcccc}
    \toprule
    % \multicolumn{2}{c}{Part}                   \\
    % \cmidrule(r){1-2}
    Marketplace     & \emph{\textbf{Products}}   & \emph{\textbf{Categories}}  & \emph{\textbf{Variety}} & \emph{\textbf{Comprehensiveness}} \\
    \midrule
    AWS Marketplace~\cite{AWS} & 3,702    &12  &\cmark &\cmark\\
    Beijing International Data Exchange~\cite{Beijing} & 714    &22 &\cmark &\cmark\\
    Canton Data Exchange~\cite{Canton} & 792   &24  &\cmark &\cmark\\
    Datarade~\cite{Datarade}     & 4,396   &25  &\cmark &\cmark\\
    Databricks~\cite{Databricks}         & 2,004          &20  &\cmark &\cmark\\ 
    Guiyang Global Big Data Exchange~\cite{Guiyang} & 669 &19  &\cmark &\cmark\\
    Snowflake~\cite{Snowflake}     & 2,585         &25  &\cmark &\cmark\\
    Shanghai Data Exchange~\cite{Shanghai} & 447  &6  &\cmark &\cmark\\
    Zhejiang Big Data Exchange~\cite{Zhejiang} & 838  &10  &\cmark &\cmark\\
        % \hline
        % Add spacing.
        \\[-3.2mm] \hline \\[-3.2mm]
    Advaneo~\cite{Advaneo} &2,376,178 &15  &\xmark &\cmark \\
    Beibu Gulf Data Exchange~\cite{Beibu} &1,016 &19 &\cmark &\xmark \\
    Carto~\cite{Carto} &11,800 &10  &\xmark &\cmark\\
    Databroker~\cite{DataBroker} &129 &10 &\xmark &\xmark \\
    Datatang~\cite{Datatang} &465 &7 &\xmark &\xmark \\
    Google Cloud~\cite{Googlecloud} &235 &30  &\cmark &\xmark \\
    Mobilithek~\cite{mobilithek} &8,015  &25  &\xmark &\cmark \\
    Nasdaq Data Link~\cite{Nasdaqdatalink} &750  &9   &\xmark &\xmark\\
    Shenzhen Data Exchange~\cite{Shenzhen} &235  &13  &\xmark &\xmark\\
    Veracity~\cite{Veracity} &186  &29  &\xmark &\cmark\\
    Western China Data Exchange~\cite{Western} &3,135 &10   &\cmark &\xmark\\
    \bottomrule
  \end{tabular}
  }
\end{table}

\subsection{Metadata Scraping} 
After identifying the candidate data marketplaces, we apply web scraping techniques to extract the required metadata. This process is non-trivial due to heterogeneity in platform design. Specifically, some websites provide complete metadata for all products in a single request, whereas others restrict access to product-level detail pages, necessitating item-by-item extraction.

\begin{figure}
    \centering
    \includegraphics[width=1\linewidth]{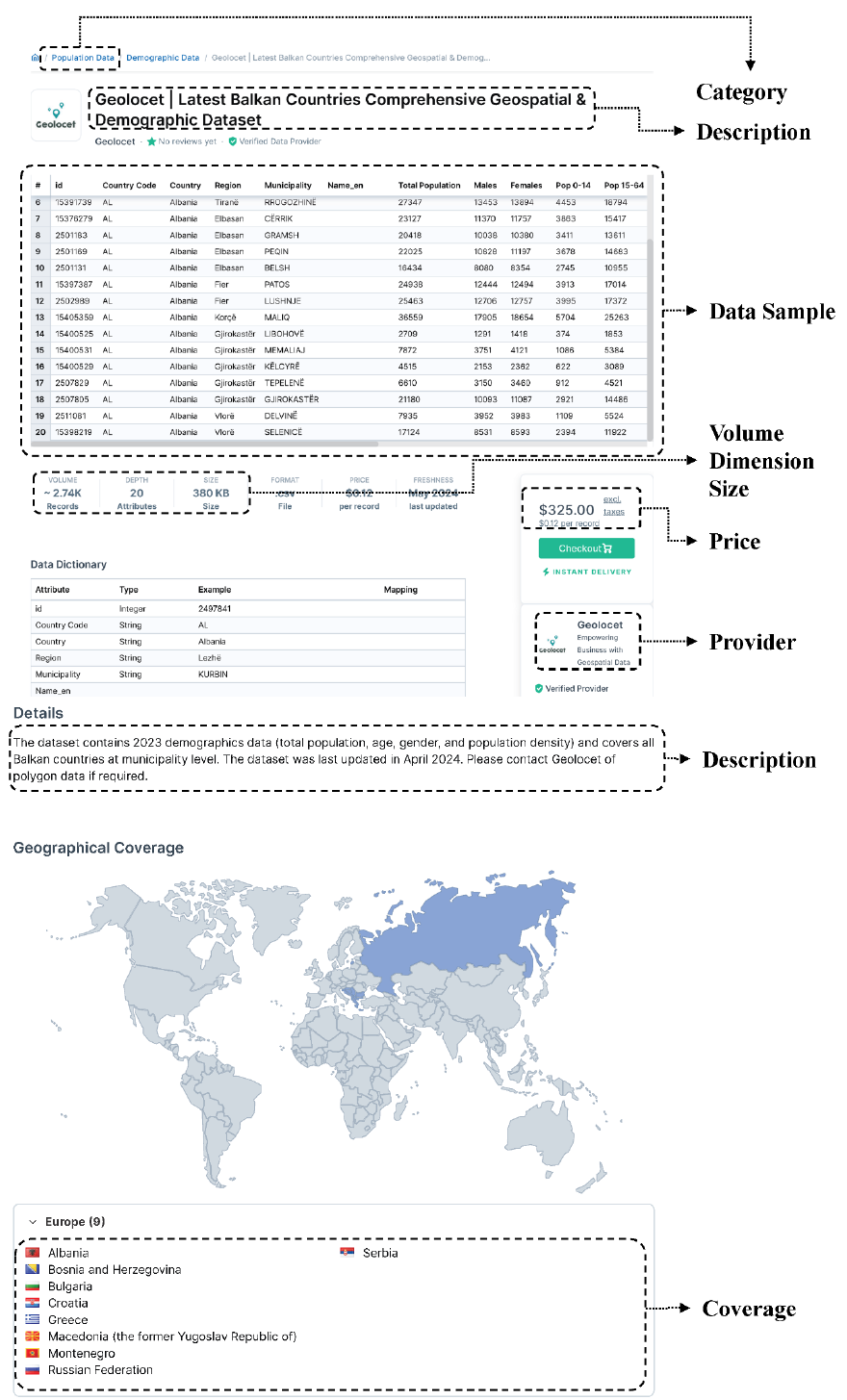}
    \caption{Scraping example of Datarade.}
    \label{fig:scrape_datarade}
\end{figure}

For the former case, we save the returned results. For the latter, we use the Selenium library \cite{Selenium} to scrape metadata for data products that are useful for data pricing.
% We scrape all useful metadata for data products that are useful for data pricing. 
Besides, we save the URL of the data product detail page for potential applications. We create XPath and CSS selector rules to precisely target and extract information by analyzing the HTML structure of the target website.  In Figure \ref{fig:scrape_datarade}, we illustrate the sources from which each metadata field is scraped. When dealing with textual fields such as \texttt{title} and \texttt{description}, we simply extract the text from the corresponding HTML elements. As for the data sample, we extract the title and rows of the sample, save this data to a file, and then store the path to the saved file in the \texttt{data\_sample} field of the corresponding data product. Once metadata is successfully extracted, it is saved in CSV format, allowing easy inspection and simplifying subsequent analysis. To avoid potential service disruptions, requests are scheduled at regular intervals, typically every three minutes. Furthermore, our crawler strictly adheres to each platform's directives and terms of service to ensure ethical and legally compliant data collection practices.

\subsection{Data Preprocessing} 

Finally, the scraped metadata is preprocessed through two primary procedures: data cleaning and data integration.

\partitle{Data Cleaning} We begin by cleaning raw metadata through three primary operations: removing duplicate items, rectifying errors, and imputing missing values. These steps lay the foundation for subsequent preprocessing while preserving dataset integrity.

\begin{enumerate}[label=\textbf{\arabic*.}, leftmargin=*, nosep]
    \item \textbf{Removing duplicate items.} Data crawling may result in the retrieval of duplicate data products. %During data crawling, duplicate data products may arise. 
    To ensure dataset uniqueness, we detect duplicates by comparing the \texttt{title} and \texttt{provider} fields. If two data products have identical fields, they are considered duplicates, and the redundant entries are removed.
    
    \item \textbf{Rectifying errors.} During the crawling process, non-UTF-8 characters may be retrieved, leading to errors when reading files containing such characters. We use a script to detect and automatically remove non-UTF-8 characters from the data.
    
    \item \textbf{Imputing missing values.} Some critical fields like \texttt{coverage} and \texttt{size} are often not explicitly provided on many data marketplaces and are only mentioned in the \texttt{description}.  To address these missing fields, we use manual annotation. Specifically, we read the \texttt{description} field to extract relevant information to fill in the missing fields. If no relevant information is available, we use default values. 
\end{enumerate}

\partitle{Data Integration} %Following data cleaning, we proceed to data integration, which aims to combine information from different sources into a unified format. 
The objective of data integration is to merge the cleaned metadata from diverse data marketplaces into a standardized schema, thereby facilitating subsequent analysis and application deployment. 
% To accommodate the varying levels of detail provided by different data marketplaces, we construct a superset of metadata designed to encompass all potentially useful attributes across data products.
Given the varying granularity of metadata, we first define a comprehensive superset that encompasses all potentially relevant attributes across data products.
% When combining metadata of data products in different data marketplaces, it is important to understand the nature of each attribute. 
During integration, it is essential to account for each attribute's characteristics. Attributes such as \texttt{title} and \texttt{provider} are independent and can be merged directly without any issues. However, when it comes to attributes like \texttt{category}, combining them directly is problematic due to the lack of a consistent pattern across different platforms. 
As shown in Table~\ref{tb:dm}, every data marketplace has its own way of classifying data products. 
For example, AWS Marketplace \cite{AWS} has 12 general categories, while Datarade \cite{Datarade} has 25 general categories and over 560 detailed classifications. Different classification standards increase the difficulty of finding similar data products across marketplaces and may result in the omission of some data products. To solve this problem, we manually map the diverse categorizations from various platforms onto the taxonomy of AWS Marketplace \cite{AWS}, ensuring consistency and creating a unified pattern to avoid the confusion that can result from a direct combination.
The AWS Marketplace classification system covers a wide range of industries and application areas, ensuring that most data products can be categorized. Additionally, as a well-established data marketplace, its classification system is widely recognized and used by a broad user base. Therefore, we choose it as our benchmark.

Besides, most of the data products in the data marketplaces are denominated in USD (\textdollar) and RMB (CN\textyen), with smaller amounts in EUR (\texteuro), GBP (\textsterling), and JPY (JP\textyen). To maintain consistency, we convert the other currencies to USD (\textdollar) using the Yearly Average Currency Exchange Rates~\cite{IRS}. Such integration eliminates redundancy and resolves conflicts, leading to a more coherent dataset. %Finally, the dataset undergoes data transformation, converting the raw dataset into a format suitable for analysis and application deployment. 

\partitle{Metadata Validation}
After integration, we perform consistency checks to reduce avoidable noise before releasing the dataset. First, we verify schema consistency by ensuring that every marketplace-specific file can be projected into the same set of standardized attributes. Second, we normalize categorical values for \texttt{price\_mode} and \texttt{update\_frequency}; for example, platform-specific phrases such as monthly billing, yearly billing, real-time updates, and on-demand updates are converted into a fixed vocabulary. Third, we validate numerical attributes by converting storage units to bytes, normalizing subscription prices to a 12-month cost, converting usage-based prices to per-use costs when the unit is disclosed, and checking that non-free products with explicit prices have positive values. Finally, for geographic coverage, we convert country abbreviations to country names and keep multi-country coverage as a list, allowing both human inspection and machine-readable filtering.

We deliberately separate automatic normalization from manual annotation. Fields that can be parsed deterministically, such as currency conversion and many update-frequency expressions, are handled by scripts. Fields that require semantic judgment, such as aligning marketplace-specific categories to the AWS taxonomy or extracting missing coverage information from long descriptions, are manually reviewed. This separation makes the construction process auditable: users can reproduce the deterministic steps from the released scripts and inspect the curated fields in the preprocessed files. Such transparency is particularly important for a pricing benchmark, since small changes in price modes, currency conversion, or category alignment can affect both downstream model performance and the interpretation of price benchmarks.

\section{Applications}
\label{section:applications}

In this section, we show that \texttt{DaDaDa} can be used for multiple downstream tasks, including data product pricing (Section~\ref{sec:pricing}), data product classification (Section~\ref{sec:classification}), and data product retrieval (Section~\ref{sec:retrieval}). We use a server with an NVIDIA A100 GPU (80GB) and 160 Intel(R) Xeon(R) Platinum 8383C CPU @ 2.70GHz for all empirical experiments in this manuscript. The operating system is Ubuntu 20.04.6.

\subsection{Data Product Pricing}
\label{sec:pricing}

\begin{figure}
    \centering
    \includegraphics[width=0.98\linewidth]{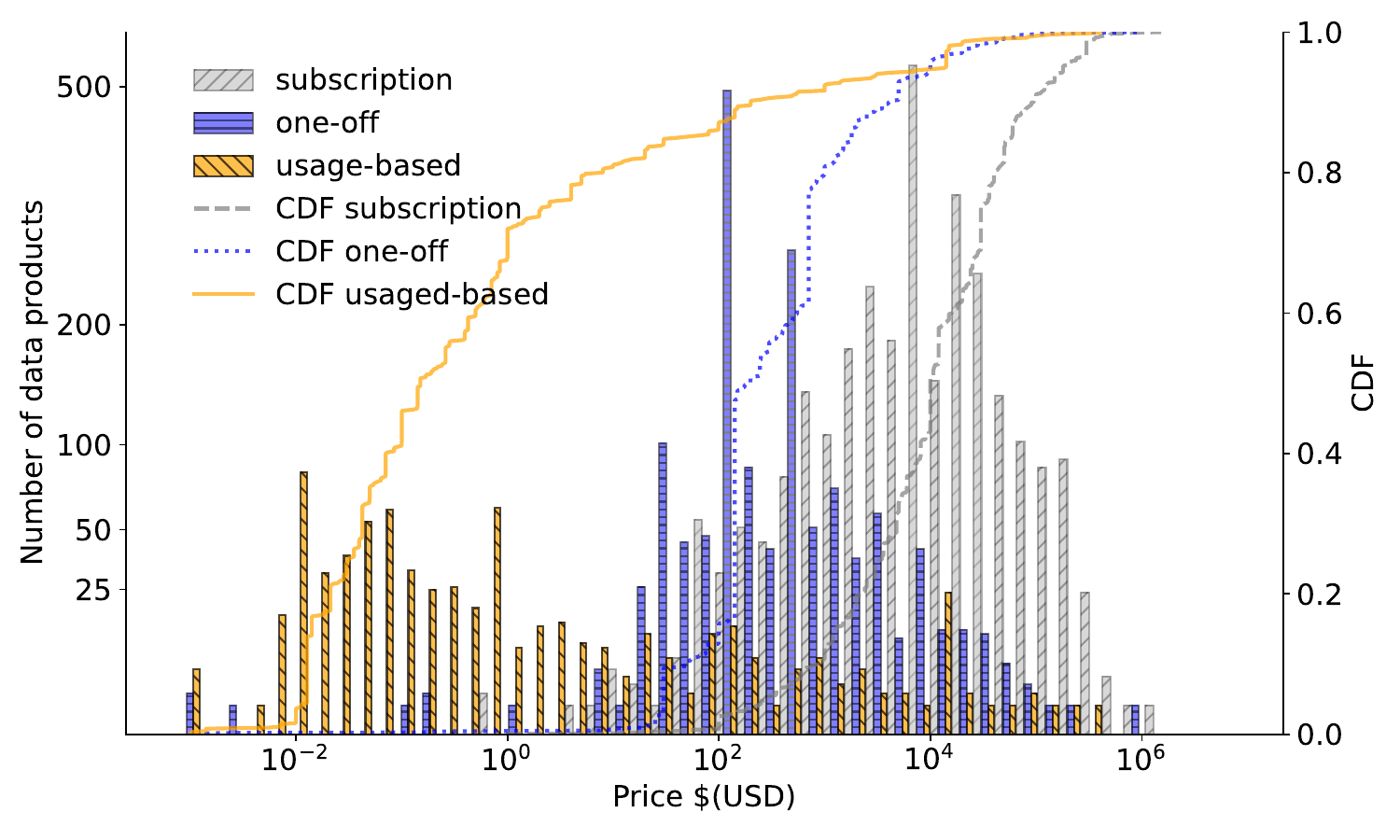}
    \caption{Histogram and CDF (Cumulative Distribution Function) of the prices in three pricing modes for data products.}
    \label{fig:price}
\end{figure}

\texttt{DaDaDa} collects important metadata and pricing information for data products, which can greatly assist in making pricing decisions. To validate the effectiveness of \texttt{DaDaDa}, we use it to train models and evaluate pricing performance. Figure \ref{fig:price} shows a histogram and the corresponding CDF of data product prices using different pricing modes. We can observe that usage-based prices are mostly concentrated between \$0.01 and \$100, one-off prices are primarily between \$10 and \$10,000, and subscription prices are mainly between \$100 and \$100,000. The prices for usage-based data are generally associated with API-type data, which is typically priced per use, resulting in lower overall prices. One-off data usually consists of static datasets, while subscription data often includes live data that is periodically updated with the latest information. This dynamic nature of subscription data makes it more valuable compared to static data, which is reflected in its higher overall price.

\noindentparagraph{Benchmark Protocol.}
We formulate data product pricing as a supervised listed-price regression task over products with explicit nonzero prices. The input consists of standardized product metadata, including semantic fields (\texttt{title} and \texttt{description}), categorical fields (\texttt{category}, \texttt{coverage}, \texttt{update\_frequency}, and \texttt{price\_mode}), and numerical fields (\texttt{volume}, \texttt{size}, and \texttt{dimension}) when available. The prediction target is the normalized USD price after applying the mode-specific standardization in Section~\ref{section:dataset_overview}. Free and negotiation-based products are excluded from this supervised regression task because their \texttt{price} values are set to zero by construction and do not represent observed listed prices. Including them would mix price regression with pricing-mode prediction and substantially change the task definition.

We predict \(\log(\texttt{price}+1)\) rather than the raw price because listed data prices span several orders of magnitude across subscription, one-off, and usage-based modes. The log transformation reduces the dominance of extremely expensive products and makes errors more comparable across price ranges. To avoid information leakage, fields that directly identify a product page, marketplace, or seller---including \texttt{url}, \texttt{platform}, \texttt{provider}, and \texttt{data\_sample}---are removed before model training. Raw \texttt{title} and \texttt{description} strings are also removed after conversion into embeddings. We evaluate all pricing models using shuffled 5-fold cross-validation with a fixed random seed and report the average \(R^2\), MAE, MSE, and standard deviation across folds. This protocol makes the task a metadata-based benchmark for estimating public listed prices rather than a memorization task over marketplaces or providers.

We preprocess the data to ensure it meets the format requirements of machine learning models. The preprocessing procedure includes the following steps. 
\begin{itemize}[leftmargin=*]
    \item \textit{Feature extraction.}  We extract features utilizing the fine-tuned XLM-RoBERTa-Large model~\cite{xlm-r-large}. Specifically, the model encodes  \texttt{description} and \texttt{title}, obtaining word embeddings from the intermediate layer (12th layer). These high-dimensional word embeddings are then reduced in size using Principal Component Analysis (PCA)~\cite{mackiewicz1993principal} to enhance computational efficiency and decrease feature count. We find the minimum number of principal components required to achieve a 95\% cumulative explained variance ratio, reducing the original 1024-dimensional word embeddings to 223 dimensions.
   \item \textit{Categorical and numerical feature processing.} For the categorical features such as \texttt{category}, \texttt{update\_frequency}, and \texttt{price\_mode}, we employ One-Hot Encoding to allow the model to interpret these data points accurately. Numerical features, including \texttt{size}, \texttt{dimension}, and \texttt{volume}, are standardized to ensure all data points are on a unified scale, thus preventing discrepancies in feature magnitudes from affecting the performance of the model.
   \item \textit{Price processing.} Considering the wide range of the target variable price, we apply a log transformation to stabilize the values and improve the robustness of the training process. This comprehensive preprocessing ensures that the data is well-prepared for the subsequent predictive modeling steps.
\end{itemize}

We select several regression models to predict the prices of data products, including random forest~\cite{breiman2001random}, $k$-nearest neighbors~\cite{kramer2011unsupervised}, gradient boosting~\cite{friedman2001greedy}, and XGBoost~\cite{chen2016xgboost}. For each of these models, we use grid search combined with cross-validation to find the best parameters and perform 5-fold cross-validation to evaluate the performance. In addition to these traditional regression models, we develop a basic deep neural network (DNN). We use ReLU~\cite{agarap2018deep} as the activation function, Adam~\cite{kingma2014adam} as the optimizer, and apply Dropout~\cite{srivastava2014dropout} to reduce overfitting.

A variety of metrics are used to evaluate the performance of models, including the coefficient of determination ($R^2$ score), mean absolute error (MAE), mean squared error (MSE), and standard deviation (SD). Table \ref{tab:prediction} presents the performance of different models on the data product pricing prediction task. The experimental results show that XGBoost consistently outperforms the other models in most evaluation metrics. 

\begin{table}[h!]
  \caption{Performance comparison of different models on the data product price prediction task. XGBoost achieves the best overall performance with the highest $R^2$ score and the lowest MSE and SD. DNN has the lowest MAE but exhibits the highest variability.}
  \label{tab:prediction}
  \centering
  \begin{tabular}{lcccc}
    \toprule               
    % \cmidrule(r){1-2}
    Model     & $R^2$ score ($\uparrow$)     & MAE ($\downarrow$) & MSE ($\downarrow$) &SD ($\downarrow$)\\
    \midrule
    $k$-Nearest Neighbors    & 0.765 & 0.808  &2.499 &1.576  \\
    DNN           &0.810  &\textbf{0.779}   &1.960   &4.368 \\
    Random Forest &0.814   & 0.876  &1.977   &1.405 \\
    Gradient Boosting     & 0.829  &0.851  &1.815 &1.346\\
    XGBoost       & \textbf{0.833}  &0.805  &\textbf{1.774}  &\textbf{1.330}\\
    \bottomrule
  \end{tabular}
\end{table} 

We also extract word embeddings from the last (24th) layer. The remaining experimental procedures are consistent with the 12th layer, where we identify the best parameters through grid search and evaluate model performance using 5-fold cross-validation. Table~\ref {tab:prediction24} summarizes the experimental results. 
XGBoost outperforms other models across all metrics, but it performs worse than the method that uses the intermediate layer as word embeddings in Section~\ref{sec:pricing}. This is because the last layer of the model is more focused on the classification task, potentially losing some information. 
\begin{table}[h]
  \caption{Performance comparison of different models on the data product price prediction task. Different from the experiment in Section \ref{sec:pricing}, we obtain word embeddings from the last layer (24th layer) instead of the intermediate layer (12th layer) of our best fine-tuned  XLM-RoBERTa-Large model. XGBoost performs the best across all metrics.}
  \label{tab:prediction24}
  \centering
  \begin{tabular}{lcccc}
    \toprule               
    % \cmidrule(r){1-2}
    Model     & $R^2$ score ($\uparrow$)     & MAE ($\downarrow$) & MSE ($\downarrow$) &SD ($\downarrow$)\\
    \midrule
    $k$-Nearest Neighbors    & 0.726 & 0.921  &2.922 &1.708  \\
    DNN           &0.745  &0.971   &2.630   &4.441 \\
    Random Forest &0.814   & 0.874  &1.977   &1.406 \\
    Gradient Boosting     & 0.808  &0.930  &2.049 &1.431\\
    XGBoost       & \textbf{0.825}  &\textbf{0.832}  &\textbf{1.860}  &\textbf{1.364}\\
    \bottomrule
  \end{tabular}
\end{table}

\noindentparagraph{Ablation Study.}
To quantify the contribution of textual and structured metadata, we conduct an ablation study using Gradient Boosting under the same 5-fold cross-validation protocol. The structured-only setting excludes text embeddings, the text-only setting uses only PCA-reduced XLM-RoBERTa-Large embeddings of \texttt{title} and \texttt{description}, and the full setting combines both feature groups. The ``Features'' column in Table~\ref{tab:pricing_ablation} reports the final input dimensionality after text PCA, categorical one-hot encoding, and numerical standardization, rather than the number of raw metadata fields.

As shown in Table~\ref{tab:pricing_ablation}, structured metadata alone provides useful pricing signals, achieving an $R^2$ score of 0.779. Text embeddings alone are less accurate, but they provide complementary semantic information: combining 12th-layer text embeddings with structured metadata improves $R^2$ to 0.834. The 12th-layer representation also outperforms the 24th-layer representation in both text-only and full settings.

\begin{table*}[t]
\begingroup
\centering
\setlength{\tabcolsep}{7pt}
\renewcommand{\arraystretch}{0.96}
\caption{Ablation study for the data product pricing task using Gradient Boosting.}
\label{tab:pricing_ablation}
\makebox[\textwidth][c]{%
\begin{tabular}{lcccccc}
\toprule
Feature setting & Embedding layer & Features & $R^2$ ($\uparrow$) & MAE ($\downarrow$) & MSE ($\downarrow$) & SD ($\downarrow$) \\
\midrule
Structured only & N/A & 279 & 0.779 & 1.047 & 2.351 & 1.533 \\
Text only & 12th & 223 & 0.652 & 1.185 & 3.716 & 1.923 \\
Text + structured & 12th & 502 & \textbf{0.834} & \textbf{0.828} & \textbf{1.763} & \textbf{1.327} \\
\midrule
Structured only & N/A & 279 & 0.779 & 1.047 & 2.351 & 1.533 \\
Text only & 24th & 55 & 0.577 & 1.389 & 4.531 & 2.125 \\
Text + structured & 24th & 334 & 0.812 & 0.906 & 2.002 & 1.414 \\
\bottomrule
\end{tabular}
}
\vspace{-6pt}
\endgroup
\end{table*}

\noindentparagraph{Pricing Results for Two Major Categories.} For this experiment, we specifically select the two major data categories in the dataset: ``Financial Services Data'' and ``Retail, Location and Marketing Data''. We train and test the pricing models separately on these two categories to assess their domain-specific performance. As shown in Table~\ref{tab:prediction_category}, the pricing model achieves higher accuracy on ``Financial Services Data'' than ``Retail, Location and Marketing Data''. This discrepancy is likely attributable to more consistent and well-defined pricing patterns in ``Financial Services Data'', which enable the model to better capture the underlying pricing logic and generate more accurate estimates. 

\begin{table}[h!]
  \caption{Performance comparison of different models on data product price prediction task, evaluated separately on two main data categories.}
  \label{tab:prediction_category}
  \centering
  \begin{tabular}{lcccc}
    \toprule  
    & \multicolumn{4}{c}{\textbf{Financial Services Data}} \\
    \cmidrule(lr){2-5}
    \textbf{Model} & $R^2$ ($\uparrow$) & MAE ($\downarrow$) & MSE ($\downarrow$) & SD ($\downarrow$) \\
    \midrule
    $k$-Nearest Neighbors & 0.818 & 0.936 & 3.056 & 1.722 \\
    DNN                   & 0.843 & 0.923 & 2.306 & 5.334 \\
    Random Forest         & 0.870 & 0.917 & 2.174 & 1.468 \\
    Gradient Boosting     & \textbf{0.882} & \textbf{0.880} & 2.174 & 1.468 \\
    XGBoost               & 0.871 & 0.903 & \textbf{2.161} & \textbf{1.466} \\
    \midrule
    & \multicolumn{4}{c}{\textbf{Retail, Location and Marketing Data}} \\
    \cmidrule(lr){2-5}
    $k$-Nearest Neighbors & 0.661 & 0.868 & 2.570 & 1.595 \\
    DNN                   & \textbf{0.801} & 0.803 & 1.604 & 3.775 \\
    Random Forest         & 0.769 & 0.848 & 1.745 & 1.316 \\
    Gradient Boosting     & 0.794 & 0.802 & \textbf{1.560} & \textbf{1.245} \\
    XGBoost               & 0.778 & \textbf{0.776} & 1.681 & 1.293 \\
    \bottomrule
  \end{tabular}
\end{table}

\noindentparagraph{Pricing Result Explanation.} The above experimental results indicate that models trained on \texttt{DaDaDa} can accurately predict data product prices. To identify the key determinants of pricing, we utilize SHAP (SHapley Additive exPlanations)~\cite{lundberg2017unified}, a game theory-based interpretability method that breaks down model predictions into contributions from individual features.

\begin{figure}
    \centering
    \includegraphics[width=0.94\linewidth]{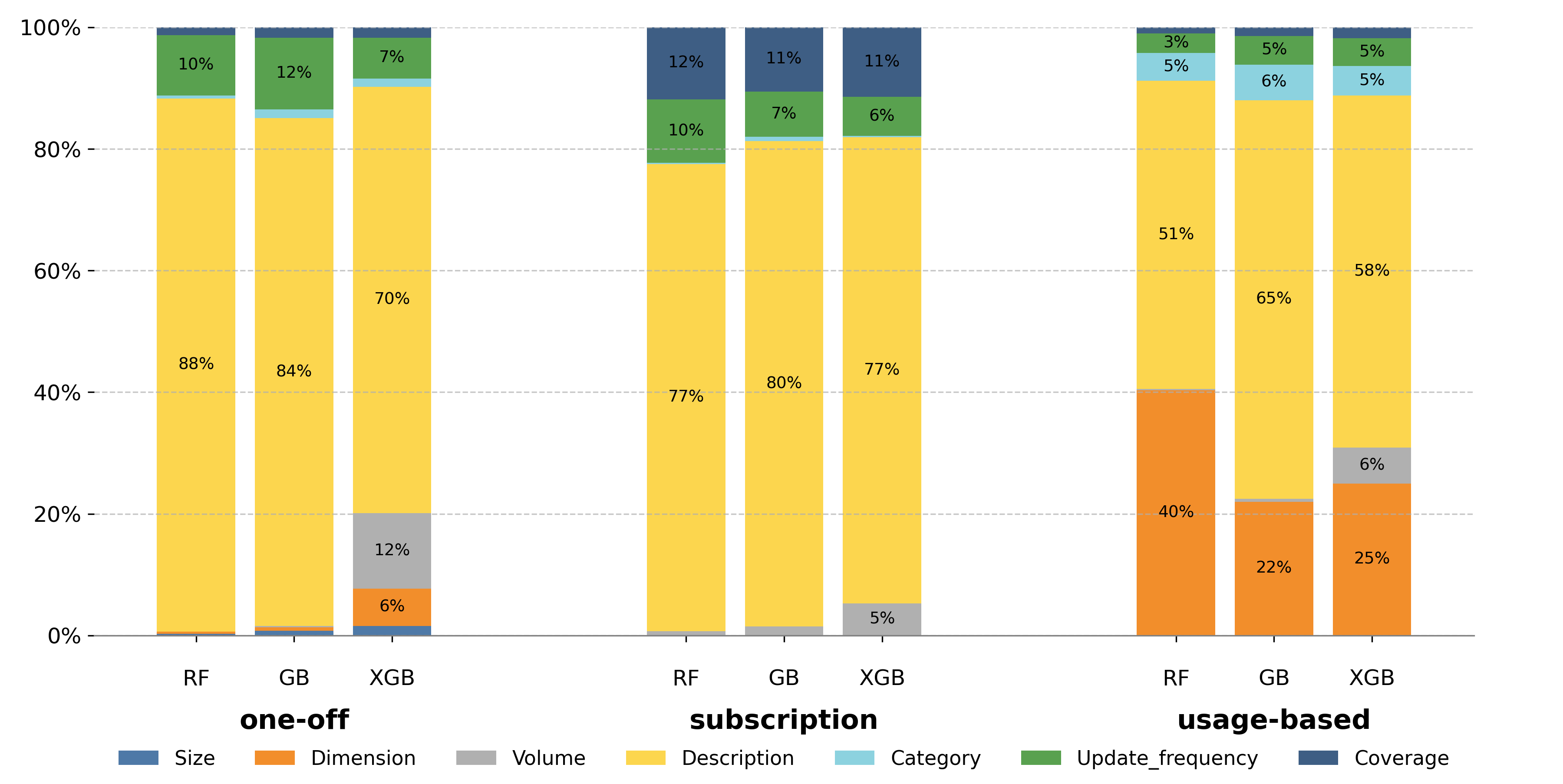}
    \caption{Results of SHAP calculation. We calculate the relative importance of different features (Size, Dimension, Volume, Description, Category, Update\_frequency, Coverage) in predicting the price of data products under three different pricing modes (one-off, subscription, usage-based), using three different machine learning models (Random Forest - RF, Gradient Boosting - GB, XGBoost - XGB).}
    \label{fig:shap}
\end{figure}

Figure~\ref{fig:price} highlights significant differences in price distributions across different pricing modes, so we conduct separate feature importance analyses for each mode using three top-performing models: Random Forest (RF), Gradient Boosting (GB), and XGBoost (XGB). As illustrated in Figure~\ref{fig:shap}, the findings are as follows:
\begin{itemize}[leftmargin=*]
\item ``Description'' is recognized as the most critical feature across all pricing modes, reflecting market realities where buyers mainly evaluate products based on descriptive content.
\item ``Update frequency'' holds significance in both one-off and subscription modes. One-off buyers appreciate frequent updates that enhance the utility of the data, while subscription users are willing to pay a premium for regularly refreshed information.
\item ``Coverage'' is crucial in the subscription mode, as subscribers need comprehensive data for long-term decision-making.
\item ``Dimension'' is particularly important in the usage-based mode, where higher-dimensional data commands premium pricing due to the increased information density offered per call. 
\item Model comparisons reveal consistent feature attention patterns, with XGBoost demonstrating the most balanced SHAP value distribution, which likely contributes to its superior pricing performance.
\end{itemize}

Table~\ref{tab:shap_summary} further summarizes the top feature groups by pricing mode. The values are grouped mean absolute SHAP scores averaged across Random Forest, Gradient Boosting, and XGBoost over the cross-validation folds. This quantitative summary confirms that \texttt{description} dominates all three pricing modes, while structured fields provide complementary mode-specific signals.

\begin{table}[h]
\begingroup
\centering
\caption{Top feature groups by grouped mean absolute SHAP value under different pricing modes.}
\label{tab:shap_summary}
\begin{tabular}{lllc}
\toprule
Pricing mode & Rank & Feature group & Mean abs. SHAP \\
\midrule
Subscription & 1 & \texttt{description} & 1.0594 \\
Subscription & 2 & \texttt{coverage} & 0.1535 \\
Subscription & 3 & \texttt{update\_frequency} & 0.1075 \\
\midrule
One-off & 1 & \texttt{description} & 1.0731 \\
One-off & 2 & \texttt{update\_frequency} & 0.1254 \\
One-off & 3 & \texttt{volume} & 0.0651 \\
\midrule
Usage-based & 1 & \texttt{description} & 1.2250 \\
Usage-based & 2 & \texttt{dimension} & 0.6059 \\
Usage-based & 3 & \texttt{category} & 0.1070 \\
\bottomrule
\end{tabular}
\endgroup
\end{table}

\noindentparagraph{Generalization.} We experimentally validate the generalization ability of our pricing model trained on the DaDaDa dataset. The DaDaDa dataset contains data products from 9 different data marketplaces. We select the data product information from 8 of these marketplaces as the training set to train our pricing model and then test the pricing prediction capability of the model on the 9th marketplace (Snowflake). This approach allows us to assess the generalization ability of the model across different data marketplaces and evaluate its performance in a new market environment. The experimental results in Table~\ref{tab:generalization} demonstrate that our model achieves strong generalization performance on the Snowflake marketplace after training on data from eight distinct marketplaces. Although data products from different marketplaces may have distinct characteristics, our model is still capable of achieving high predictive accuracy in an unseen marketplace.
\begin{table}[h]
  \caption{Generalization performance comparison of different models on the data product price prediction task. The model is trained on eight source marketplaces and evaluated on the held-out Snowflake marketplace, demonstrating generalization capability in new marketplace environments.}
  \label{tab:generalization}
  \centering
  \begin{tabular}{lcccc}
    \toprule               
    % \cmidrule(r){1-2}
    Model     & $R^2$ score ($\uparrow$)     & MAE ($\downarrow$) & MSE ($\downarrow$) &SD ($\downarrow$)\\
    \midrule
    Random Forest &0.675   & 1.733  &4.327   &1.821 \\
    Gradient Boosting     & 0.741  &\textbf{1.514}  &\textbf{3.445} &1.702\\
    XGBoost       & \textbf{0.744}  &1.887  &5.350  &\textbf{1.594}\\
    \bottomrule
  \end{tabular}
\end{table}

\subsection{Data Product Classification}
\label{sec:classification}
\texttt{DaDaDa} supports training classification models to assist in classifying new data products into predefined categories based on their metadata. To validate its effectiveness, we fine-tune two multilingual pretrained models: mBERT (bert-base-multilingual-cased)~\cite{mbert} and XLM-RoBERTa-Large~\cite{xlm-r-large}, testing their classification performance.

We divide \texttt{DaDaDa} into three parts: a training set (60\%), a validation set (20\%), and a test set (20\%).   
The category of a data product can be inferred from its \texttt{title} and \texttt{description}. These two fields are mainly in English and Chinese, with a small portion in Japanese.  To work with this multilingual data, we choose mBERT and XLM-RoBERTa-Large, which have been pretrained in multiple languages. We concatenate \texttt{title} and \texttt{description}, and input them into the models for fine-tuning. We assess the performance using the validation set and select the best fine-tuned model after multiple rounds of training. Finally, we evaluate the performance of the best fine-tuned model using the test set. 

Table \ref{tab:class_result} presents the performance of the fine-tuned models on the test set. XLM-RoBERTa-Large shows better classification performance in most categories, outperforming mBERT overall. Both models achieve an average F1-score of over 0.8, demonstrating that our dataset can effectively aid in classifying data products and promote the standardization of classification.  Table~\ref{tab:confusion_matrix} reports the corresponding confusion matrix.

%We select the best fine-tuned model on the validation set to test its performance on the test set.
\begin{table*}\centering
\small
\caption{{\color{black} Performance of mBERT and XLM-RoBERTa-Large on different data product categories.  The XLM-RoBERTa-Large model has better classification results in most categories.}}\label{tab:class_result}

{
\renewcommand{\arraystretch}{1.02}
\setlength{\tabcolsep}{5.8pt}
\begin{tabular}{lcccccccc}
\toprule
{\color{black}} \multirow{2}{*}{Category}& \multicolumn{4}{c}{mBERT} & \multicolumn{4}{c}{XLM-RoBERTa-Large}\\
\cmidrule(l{2pt}r{2pt}){2-5} \cmidrule(l{2pt}r{2pt}){6-9}
{\color{black}} & Precision & Recall & F1 & Mean & Precision & Recall & F1 & Mean\\
\midrule
Automotive &0.667 &\textbf{0.701} &0.684 &0.684 &\textbf{0.716} &0.688 &\textbf{0.702} &\textbf{0.702}\\
Environmental &0.774 &\textbf{0.739} &\textbf{0.756} &\textbf{0.756} &\textbf{0.782} &0.693 &0.735 &0.737\\
Financial Services &0.839 &0.809 &0.824 &0.824 &\textbf{0.840} &\textbf{0.851} &\textbf{0.846} &\textbf{0.846}\\
Gaming &1.000 &0.500 &0.667 &0.722 &\textbf{1.000} &\textbf{1.000} &\textbf{1.000} &\textbf{1.000}\\
Healthcare and Life Sciences &0.864 &0.851 &0.858 &0.858 &\textbf{0.878} &\textbf{0.851} &\textbf{0.864} &\textbf{0.864}\\
Manufacturing &\textbf{0.927} &\textbf{0.927} &\textbf{0.927} &\textbf{0.927} &0.906 &0.921 &0.914 &0.914\\
Media and Entertainment &0.739 &\textbf{0.790} &\textbf{0.764} &\textbf{0.764} &\textbf{0.756} &0.749 &0.752 &0.752\\
Public Sector &\textbf{0.743} &\textbf{0.774} &\textbf{0.758} &\textbf{0.758} &0.732 &0.770 &0.751 &0.751\\
Resources &0.634 &0.607 &0.620 &0.620 &\textbf{0.717} &\textbf{0.650} &\textbf{0.682} &\textbf{0.683}\\
Retail, Location and Marketing &0.848 &0.854 &0.851 &0.851 &\textbf{0.849} &\textbf{0.878} &\textbf{0.863} &\textbf{0.863}\\
Telecommunications &0.700 &\textbf{0.714} &0.707 &0.707 &\textbf{0.748} &0.671 &\textbf{0.707} &\textbf{0.709}\\
Other &0.562 &0.474 &0.514 &0.517 &\textbf{0.733} &\textbf{0.579} &\textbf{0.647} &\textbf{0.653}\\
\midrule
Weighted Average &0.810 &0.809 &0.809 &0.809 &\textbf{0.820} &\textbf{0.821} &\textbf{0.820} &\textbf{0.820}\\
\bottomrule
\end{tabular}}

\end{table*}%

\begin{table*}[htbp]
\centering
\caption{Confusion Matrix for XLM-RoBERTa-Large Classification.}
\label{tab:confusion_matrix}

\resizebox{0.9\textwidth}{!}{
\scriptsize
\begin{tabular}{lcccccccccccc}
\toprule
 & \multicolumn{12}{c}{\textbf{Predicted Class}} \\
\cmidrule(lr){2-13}
\textbf{True Class} & \textbf{Auto.} & \textbf{Env.} & \textbf{Fin.} & \textbf{Gam.} & \textbf{Health} & \textbf{Manuf.} & \textbf{Media} & \textbf{Other} & \textbf{Public} & \textbf{Res.} & \textbf{Retail} & \textbf{Tele.} \\
\midrule
Automotive & 53 & 2 & 5 & 0 & 0 & 2 & 1 & 0 & 4 & 0 & 8 & 2 \\
Environmental & 1 & 61 & 6 & 0 & 1 & 0 & 2 & 0 & 6 & 8 & 2 & 1 \\
Financial & 3 & 5 & 600 & 0 & 3 & 4 & 15 & 0 & 3 & 3 & 60 & 9 \\
Gaming & 0 & 0 & 0 & 2 & 0 & 0 & 0 & 0 & 0 & 0 & 0 & 0 \\
Healthcare & 1 & 1 & 4 & 0 & 172 & 1 & 3 & 0 & 4 & 3 & 12 & 1 \\
Manufacturing & 3 & 1 & 1 & 0 & 3 & 164 & 0 & 0 & 3 & 0 & 1 & 2 \\
Media & 6 & 0 & 6 & 0 & 2 & 0 & 164 & 0 & 6 & 0 & 29 & 6 \\
Other & 0 & 0 & 0 & 0 & 0 & 0 & 2 & 11 & 0 & 0 & 3 & 3 \\
Public Sector & 2 & 1 & 16 & 0 & 8 & 3 & 4 & 0 & 167 & 7 & 6 & 3 \\
Resources & 1 & 3 & 8 & 0 & 1 & 0 & 0 & 0 & 12 & 76 & 12 & 4 \\
Retail & 4 & 4 & 55 & 0 & 5 & 4 & 19 & 3 & 15 & 6 & 1013 & 26 \\
Telecomm & 0 & 0 & 13 & 0 & 1 & 3 & 7 & 1 & 8 & 3 & 47 & 169 \\
\bottomrule
\end{tabular}}
\end{table*}

\subsection{Data Product Retrieval}
\label{sec:retrieval}
\begin{figure}[h]
    \centering
    \includegraphics[width=0.97\linewidth]{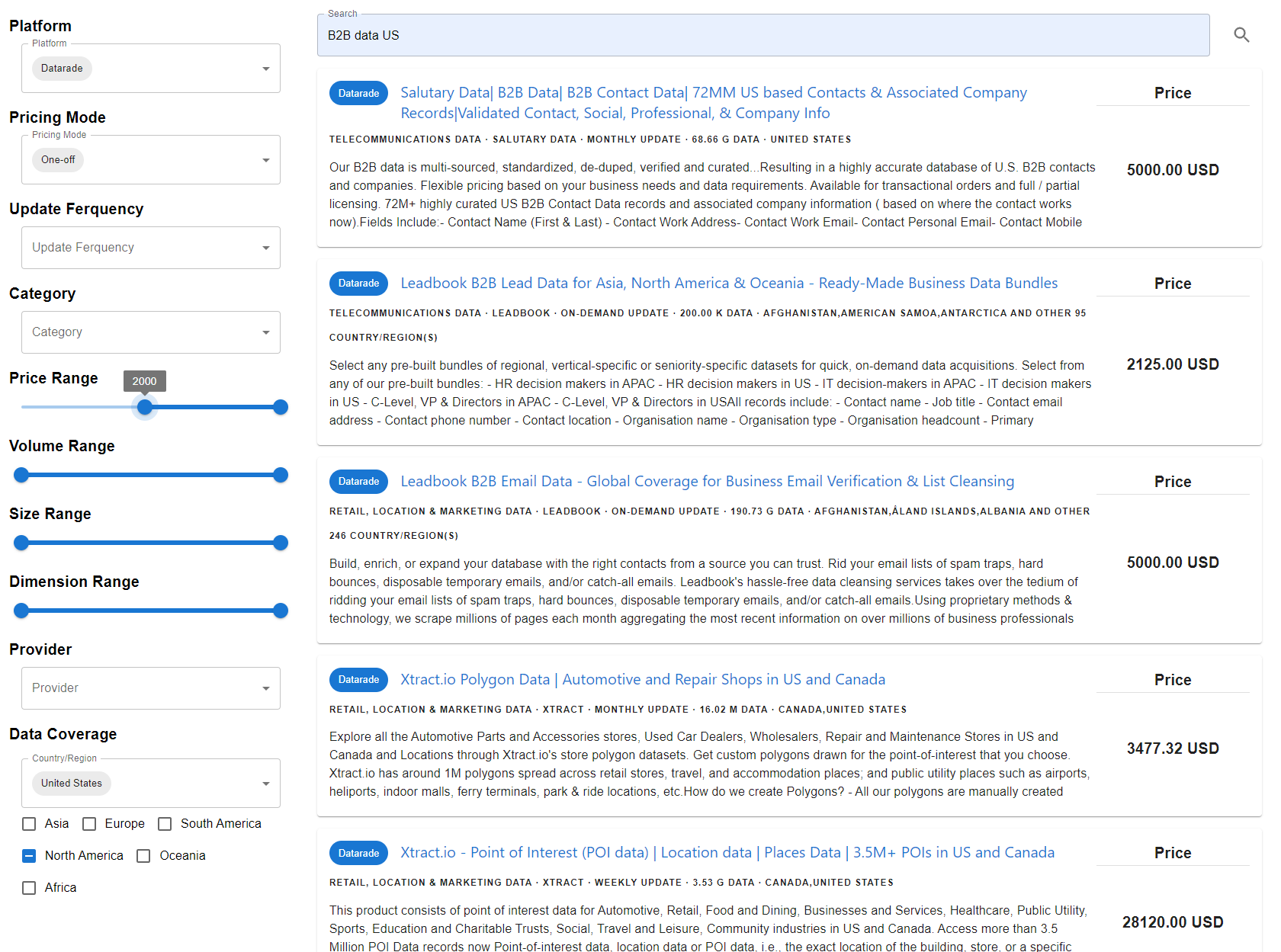}
    \caption{A screenshot of the data product search engine. }
    \label{fig:search}
\end{figure}

When searching for a suitable dataset for machine learning or other purposes, it can be frustrating to sift through many data marketplaces and thousands of data products, only to find that very few meet the requirements for training. While general search engines are the first thing that comes to mind for online searches~\cite{DBLP:conf/icde/FernandezAKYMS18}, they are not well-equipped to handle the specific task of searching for data products. This is due to the fact that different data marketplaces offer different types of data products with varying levels of detail in the metadata, making it a daunting task to gather comparable results and decide which product to choose. \texttt{DaDaDa} makes it possible to create a vertical search engine for data products collected from major open-access data marketplaces around the world. To demonstrate how \texttt{DaDaDa} can streamline the data search process, we use it to create a search engine based on Elasticsearch.

Figure \ref{fig:search} displays the search engine based on \texttt{DaDaDa}. The left part of the figure shows the metadata filter, where users can adjust settings to obtain filtered results. The available options in \texttt{DaDaDa} can be categorized into two types of filters: range filters and keyword filters. Price, volume, size, and dimension are examples of range filters, allowing users to set minimum and maximum values for each option. The remaining features are keyword filters, offering multiple options for users. The system generates a query based on the filter parameters and the search text in the input box at the top of the page. The most relevant query results are displayed below the input box, along with necessary metadata such as title, price, platform, and part of the description. Users can click the title to navigate to the detail page of the product for more information or to make a purchase.

For example, to find data products that contain B2B data covering the United States, the user can simply input ``B2B data US'', and all results matching the search text are displayed below. If the user has more detailed requirements, such as products from Datarade with a one-off pricing mode and a price higher than 2000 USD, the user can set the ``Platform'', ``Pricing Mode'', ``Price Range'', and ``Data Coverage'' filters accordingly for a refined result.

\noindentparagraph{Retrieval Implementation.}
The search engine indexes each data product as a document containing textual fields, structured categorical fields, and numerical fields. The \texttt{title}, \texttt{description}, and geographic coverage fields are indexed for full-text matching, while \texttt{platform}, \texttt{provider}, \texttt{category}, \texttt{price\_mode}, and \texttt{update\_frequency} are represented as keyword filters. Numerical attributes such as \texttt{price}, \texttt{volume}, \texttt{size}, and \texttt{dimension} are represented as range-query fields. This mixed indexing design is useful for data-product search because user intent is usually both semantic and constrained: a buyer may search for ``mobility data'', but also require a particular region, a maximum price, or a specific pricing model.

We additionally include a country synonym analyzer for geographic coverage. This helps bridge query variations such as country names, abbreviations, and region-related terms, which frequently appear in marketplace descriptions. Because every retrieved item links back to its original marketplace URL, the search engine acts as a cross-marketplace discovery layer rather than replacing the original transaction platforms. The implementation demonstrates that \texttt{DaDaDa} can support not only offline model training, but also user-facing data discovery systems where pricing information, metadata filters, and original product pages are presented together.

\noindentparagraph{Retrieval Case Study.}
The Elasticsearch prototype maps each request to three query channels: full-text matching over \texttt{title}, \texttt{desc}, and \texttt{geo\_coverage}; keyword filtering over \texttt{platform}, \texttt{provider}, \texttt{category}, \texttt{pricing\_type}, and \texttt{freq}; and range filtering over \texttt{price}, \texttt{volume}, \texttt{size}, and \texttt{dimension}. These channels mirror the interface in Figure~\ref{fig:search}: buyers first express a semantic need, such as ``B2B data US'' or ``mobility data Germany'', and then narrow results by marketplace, geography, billing model, update frequency, or numerical constraints. The mapping also keeps the case reproducible because each interface constraint corresponds to a field in the released index schema.

For example, if a buyer needs B2B data covering the United States from Datarade with one-off pricing above 2,000 USD, the prototype matches the query text, uses country synonyms to align ``US'', ``U.S.'', and ``United States'', and applies \texttt{platform}, \texttt{pricing\_type}, and \texttt{price} filters. Returned product cards show comparable metadata, including marketplace, title, provider, category, update frequency, coverage, price or free/negotiation status, a short description, and the original URL. This case study does not claim end-user retrieval effectiveness; rather, it shows that the unified \texttt{DaDaDa} schema is sufficient to build a practical vertical search layer over heterogeneous data marketplaces.

The example also exercises all three field families in the released schema, making it useful for checking whether text analyzers, keyword filters, and numerical range predicates work together. Since each result links back to the original product page, the prototype remains a discovery layer that helps buyers compare candidates before inspecting licensing and purchase details on the source marketplace.

\section{Availability, Ethics, and Limitations}
\label{section:availability_ethics_limitations}

\noindentparagraph{Availability and Reproducibility.}
We release \texttt{DaDaDa} through the project repository at \url{https://github.com/ZJU-DIVER/DaDaDa}, together with a Kaggle mirror for convenient dataset access. The repository contains the final standardized dataset, raw crawled files, intermediate preprocessed files, sample data files, crawling code, preprocessing scripts, experiment scripts, and metadata documentation. The pricing pipeline includes scripts for extracting XLM-RoBERTa-Large embeddings, applying PCA, constructing regression features, training pricing models, and computing SHAP-based feature importance. The classification pipeline includes model fine-tuning and test scripts for multilingual pretrained models. The retrieval pipeline includes the Elasticsearch index schema and a frontend demonstration. Releasing these components together allows researchers to reproduce our task results and to extend the benchmark with new pricing models, classification methods, or retrieval interfaces.

\noindentparagraph{Ethical Considerations.}
The dataset is constructed from publicly accessible product metadata rather than private transaction records or raw customer data. We collect product-level descriptions, public seller-provided attributes, prices, and URLs that are visible on data marketplace pages. During crawling, requests are scheduled at controlled intervals and platform directives are respected to avoid service disruption. We do not redistribute private buyer information, transaction histories, or any non-public information. When product samples are publicly provided by the source marketplace, we store only the sample files as released by the marketplace; otherwise, the \texttt{data\_sample} field records that no sample is available.

Because the dataset describes commercial data products, it should be used as a benchmark for research, marketplace analysis, and decision support rather than as a tool for copying or reselling products. Prices and descriptions remain attributable to their original marketplaces through the public URLs. Users should check the original product page and licensing terms before purchasing, using, or redistributing any referenced data product. This is especially important for sensitive domains such as healthcare, finance, mobility, and personal-data-related products, where marketplace listings may involve additional legal or contractual constraints.

\noindentparagraph{Limitations.}
\texttt{DaDaDa} is a versioned snapshot of public marketplace listings, so it is intended as a benchmark for listed-price estimation and metadata-driven comparison under transparent and reproducible assumptions rather than a record of realized transactions. Public pages may change over time, and private discounts, enterprise contracts, bundling, or custom licensing terms are usually not visible. Some structured fields are also disclosed unevenly across platforms; we preserve such products to maintain broad marketplace coverage and encode unavailable numeric values consistently instead of inventing unsupported properties. Negotiation-based products remain useful for marketplace and retrieval analysis even when they are not included in listed-price regression. These design choices keep the dataset aligned with how buyers observe data products in practice, while the released URLs, crawlers, preprocessing scripts, and unified schema allow future snapshots to refresh listings and compare new methods under the same public-data protocol.

\section{Conclusion}
\label{section:conclusion}

% \todo{To address these problems, We collect xxx items from xxx prevalent data marketplace from China and America}

% In this paper, We present \texttt{DaDaDa}, a comprehensive dataset of data products collected from major data marketplaces. It addresses the gap in data product pricing by including various features that describe and drive the price of a data product. We use this dataset to fine-tune two pretrained multilingual models that classify data products into unified categories based on their title and description. Additionally, to demonstrate how this dataset can improve the data retrieval process, we develop a data product search engine. Furthermore, we train a pricing model to predict the price of data products based on their metadata. Experimental results show that the categorization model and pricing model achieve high accuracy.
In this paper, we introduce \texttt{DaDaDa}, the first comprehensive dataset for data product pricing. This dataset fills the gap in data product pricing by including various features that describe and influence the price of a data product. Additionally, we develop a data product search engine to improve the data retrieval process. Experimental results show that both the pricing and classification tasks achieve high accuracy, highlighting the effectiveness of \texttt{DaDaDa} in enhancing data product pricing and classification. 
Beyond these tasks, \texttt{DaDaDa} provides a reusable benchmark for empirical data-marketplace research. The released metadata, URLs, crawlers, preprocessing scripts, and task code allow future work to refresh public listings and compare pricing, classification, and retrieval methods under a shared protocol. This gives researchers a common reference point for studying data products across marketplaces.

The importance of data markets is increasingly recognized and the field continues to develop rapidly. In contributing to this evolving landscape, we plan to develop an online metadata crawling system that can automatically gather the latest product metadata and process it into a standardized format to ensure regular updates for the dataset. Such temporal extensions will make it possible to study how sellers revise prices and how marketplace-level disclosure practices evolve.

% Using \texttt{DaDaDa}, we fine-tune two pretrained multilingual models to classify data products into unified categories based on their titles and descriptions. To demonstrate its potential in improving the data retrieval process, we develop a data product search engine. Furthermore, we train a pricing model to predict the price of data products based on their metadata.

% \noindentparagraph{Limitations.} The main limitation is that \texttt{DaDaDa} provides the current prices and features of data products, while these prices can fluctuate over time due to market factors. Data sellers may dynamically adjust product prices based on metrics such as page views and transaction volumes to maximize revenue. However, this information is typically not visible to regular users and cannot be obtained through a single web scraping, which could affect the accurate assessment of data product prices.

\newpage
\bibliographystyle{abbrv}
\IEEEtriggeratref{37}
\bibliography{dataset}

%%%%%%%%%%%%%%%%%%%%%%%%%%%%%%%%%%%%%%%%%%%%%%%%%%%%%%%%%%%%
% \newpage
% \appendix

% \section{Appendix}
% \label{appendix}

% \subsection{Ethics Statement}
% The development and dissemination of the \texttt{DaDaDa} dataset adhere to stringent ethical standards to ensure the integrity of the data, and the responsible use of the information. This transparency ensures that users of the \texttt{DaDaDa} dataset understand the origin of the data and the context in which it was collected. 

% \subsection{Disclosure of AI Assistance}
% This work employed large language models to polish the English writing and to assist with a small portion of code implementation. All core contributions, including problem formulation, methodology design, experimental analysis, and interpretation of results, were conceived and executed solely by the authors.

% \noindentparagraph{Limitations} Data pricing is a vast and fast-growing field, and there are important tasks and datasets yet to be included in \texttt{DaDaDa}. However, \texttt{DaDaDa} is an ongoing effort and we strive to continuously include more datasets and tasks in the future. 

\end{document}